\documentclass[journal]{IEEEtran}

\usepackage[cmex10]{amsmath}
\usepackage{amssymb}
\usepackage{bbm}
\usepackage{epsfig}
\usepackage{cite}
\def\citet{\cite}
\def\citep{\cite}
\def\R{\mathbbm{R}}
\def\equaldef{\overset{\underset{\mathrm{def}}{}}{=}}

\begin{document}
% paper title
% can use linebreaks \\ within to get better formatting as desired
\title{Efficient EM Training of Gaussian Mixtures with Missing Data}
%
%
% author names and IEEE memberships
% note positions of commas and nonbreaking spaces ( ~ ) LaTeX will not break
% a structure at a ~ so this keeps an author's name from being broken across
% two lines.
% use \thanks{} to gain access to the first footnote area
% a separate \thanks must be used for each paragraph as LaTeX2e's \thanks
% was not built to handle multiple paragraphs
%

\author{Olivier~Delalleau,
        Aaron~Courville,
        and~Yoshua~Bengio% <-this % stops a space
\thanks{Olivier Delalleau, Aaron Courville and Yoshua Bengio
are with the Department of
Computer Science and Operations Research, University of Montreal, Canada.}%
%\thanks{Manuscript received April 19, 2005; revised January 11, 2007.}%
}

\markboth{}%
{Delalleau \MakeLowercase{\textit{et al.}}: Efficient EM Training of Gaussian Mixtures with Missing Data}

%\markboth{Journal of \LaTeX\ Class Files,~Vol.~6, No.~1, January~2007}%
%{Shell \MakeLowercase{\textit{et al.}}: Bare Demo of IEEEtran.cls for Journals}
% The only time the second header will appear is for the odd numbered pages
% after the title page when using the twoside option.
% 
% *** Note that you probably will NOT want to include the author's ***
% *** name in the headers of peer review papers.                   ***
% You can use \ifCLASSOPTIONpeerreview for conditional compilation here if
% you desire.

% If you want to put a publisher's ID mark on the page you can do it like
% this:
%\IEEEpubid{0000--0000/00\$00.00~\copyright~2007 IEEE}
% Remember, if you use this you must call \IEEEpubidadjcol in the second
% column for its text to clear the IEEEpubid mark.

% make the title area
\maketitle

\begin{abstract}
In data-mining applications, we are frequently faced with a large
fraction of missing entries in the data matrix, which is problematic for
most discriminant machine learning algorithms. A solution that we explore
in this paper is the use of a generative model (a mixture of Gaussians) to compute the conditional
expectation of the missing variables given the observed
variables.
Since training a Gaussian
mixture with many different patterns of missing values can be
computationally very expensive, we introduce a spanning-tree based
algorithm that significantly speeds up training
in these conditions.
We also observe that good results can be obtained by 
using the generative model to fill-in the missing values for a
separate discriminant learning algorithm.
\end{abstract}

% Note that keywords are not normally used for peerreview papers.
\begin{IEEEkeywords}
Gaussian mixtures, missing data, EM algorithm, imputation.
\end{IEEEkeywords}

% For peer review papers, you can put extra information on the cover
% page as needed:
% \ifCLASSOPTIONpeerreview
% \begin{center} \bfseries EDICS Category: 3-BBND \end{center}
% \fi
%
% For peerreview papers, this IEEEtran command inserts a page break and
% creates the second title. It will be ignored for other modes.
\IEEEpeerreviewmaketitle

\section{Introduction}
% The very first letter is a 2 line initial drop letter followed
% by the rest of the first word in caps.
% 
% form to use if the first word consists of a single letter:
% \IEEEPARstart{A}{demo} file is ....
% 
% form to use if you need the single drop letter followed by
% normal text (unknown if ever used by IEEE):
% \IEEEPARstart{A}{}demo file is ....
% 
% Some journals put the first two words in caps:
% \IEEEPARstart{T}{his demo} file is ....
% 
% Here we have the typical use of a "T" for an initial drop letter
% and "HIS" in caps to complete the first word.
%\IEEEPARstart{T}{his} demo file is intended to serve as a ``starter file''
%for IEEE journal papers produced under \LaTeX\ using
%IEEEtran.cls version 1.7 and later.
% You must have at least 2 lines in the paragraph with the drop letter
% (should never be an issue)
%I wish you the best of success.

\IEEEPARstart{T}{he} presence of missing values in a dataset often makes it difficult to
apply a large class of machine learning algorithms.  In many real-world
data-mining problems, databases may contain missing values due directly to
the way the data is obtained (e.g. survey or climate data), or also
frequently because the gathering process changes over time (e.g. addition
of new variables or merging with other databases).  One of the simplest
ways to deal with missing data is to discard samples and/or variables that
contain missing values.  However, such a technique is not suited to
datasets with many missing values; these are the focus of this paper. 

Here, we propose to use a generative model (a mixture of Gaussians with full covariances) to learn
the underlying data distribution and replace missing values by their
conditional expectation given the observed variables.
A mixture of Gaussians is particularly well suited to generic data-mining problems because:
\begin{itemize}
\item By varying the number of mixture components, one can combine the advantages
of simple multivariate parametric models (e.g. a single Gaussian), that
usually provide good generalization and stability properties, to those of
non-parametric density estimators (e.g. Parzen windows, that puts one Gaussian
per training sample~\citep{Parzen62}), that avoid making strong assumptions on the
underlying data distribution.
\item The Expectation-Maximization (EM) training algorithm~\citep{Dempster77}
naturally handles missing values and provides the missing values imputation
mechanism.
One should keep in mind that the EM algorithm assumes the missing data
are ``Missing At Random'' (MAR), i.e. that the probability of
variables to be missing does not depend on the actual value of missing variables.
Even though this assumption will not always hold in practical applications like
those mentioned above, we note that applying the EM algorithm might still yield sensible
results in the absence of theoretical justifications.
\item Training a mixture of Gaussians scales only linearly with the number of samples, which is attractive for
large datasets (at least in low dimension, and we will see in this paper
how large-dimensional problems can be tackled).
\item Computations with Gaussians often lead to analytical solutions that
avoid the use of approximate or sampling methods.
\item When confronted with the supervised problem of learning
a function $y = f(x)$, a mixture of Gaussians trained to learn a joint
density $p(x,y)$ directly provides a least-square estimate of $f(x)$ by
$\hat{y} = E[Y | x]$.
\end{itemize}

Even though such mixtures of Gaussians can indeed be applied directly to
supervised problems~\citep{Zoubin-nips94}, we will see in experiments
that using them for missing value imputation before
applying a discriminant learning algorithm yields better results.
This observation is in line with the common belief that generative
models that are trained to learn the global data distribution are
not directly competitive with discriminant algorithms for prediction tasks~\citep{Bahl86}.
However, they can provide useful information regarding the data
that will help such discriminant algorithms to reach better accuracy.

The contributions in this paper are two-fold:
\begin{enumerate}
\item We explain why the basic EM training algorithm is not practical in large-dimensional
applications in the presence of missing values, and we
propose a novel training algorithm that significantly speeds it up.
\item We show, both by visual inspection on image data and by
for imputed values for classification,
%feeding the imputed values to a classification algorithm,
how a mixture of Gaussians can model the data distribution so
as to provide a valuable tool for missing values imputation.
\end{enumerate}
Note that an extensive study of Gaussian mixture training
and missing value imputation algorithms is
out of the scope of this paper.
Various variants of EM have been proposed in the past (e.g.~\citep{Lin+al-2006}),
while we focus here on the ``original'' EM, showing how it can be solved
{\bf exactly} at a significantly lower computational cost.
For missing value imputation, statisticians may prefer to
draw from the conditional distribution instead of inputing its mean,
as the former better preserves data covariance~\citep{DiZio+al-2007}.
In machine learning, the fact that a value is missing may also be by itself
a useful piece of information worth taking into account (for
instance by adding extra binary inputs to the model, indicating whether
each value was observed).
All these are important considerations that one should keep in mind,
but they will not be addressed here.

% Remark: these are OLD todos.
% TODO Can we do semi-supervised learning, i.e. missing values imputation
% while using the validation and test sets.
% TODO Talk about regularization somewhere?

\iffalse
In the following section, we present the standard EM algorithm for training
mixtures of Gaussians in the presence of missing data.  The fast EM
algorithm is detailed in section~\ref{sec:scaling}, while
section~\ref{sec:experiments} describes experimental results assessing the
efficiency of the proposed method for missing value imputation.  Finally,
section~\ref{sec:conclusion} concludes this paper and discusses several
possible extensions.
\fi

\section{EM for Gaussian Mixtures with Missing Data}
\label{sec:em}

% Describe the standard algorithm.
% Talk about missing values imputation.

In this section, we present the EM algorithm for learning a mixture of
Gaussians on a dataset with missing values.  The notations we use are as
follows.  The training set is denoted by ${\cal D} = \{x^1, \ldots, x^n\}$.
Each sample $x^i \in \R^d$ may have different missing variables. A {\em
  missing pattern} is a maximal set of variables that are simultaneously missing in
at least one training sample.  When the input dimension $d$ is high, there
may be many different missing patterns, possibly on the same order as the
number of samples $n$ (since the number of possible missing patterns is $2^d$).
For a sample $x^i$, we denote by $x^i_o$ and
$x^i_m$ the vectors corresponding to respectively the observed and missing
variables in $x^i$.  Similarly, given $x^i$, a symmetric $(d \times d)$
matrix $M$ can be split into four parts corresponding to the observed and
missing variables in $x^i$, as follows:
\begin{itemize}
\item $M_{oo}$ contains elements $M_{kl}$ where variables $k$ and $l$
are observed in $x^i$,
\item $M_{mm}$ contains elements $M_{kl}$ where both variables $k$ and $l$
are missing in $x^i$,
\item $M_{om} = M_{mo}^T$ contains elements $M_{kl}$ where variable $k$ is
observed in $x^i$, while variable $l$ is missing.
\end{itemize}
It
is important to keep in mind that with these notations, we have for instance
$M_{mm}^{-1} \equaldef (M_{mm})^{-1} \neq (M^{-1})_{mm}$, i.e. the inverse of a
sub-matrix is not the sub-matrix of the inverse.
Also, although to keep notations simple we always write for instance $M_{oo}$,
the observed part depends on the sample currently being
considered: for two different samples, the $M_{oo}$ matrix may represent
a different sub-part of $M$.
It should be clear from the context which sample is being considered.

The EM algorithm \citep{Dempster77} can be directly applied in the presence
of missing values \citep{Little+Rubin-2002}.  As shown in~\citet{Zoubin-nips94}, for a mixture of Gaussians
the computations for the two
steps (Expectation and Maximization) of the algorithm are\footnote{Although
these equations assume constant (and equal) mixing weights, they can trivially be extended
to optimize those weights as well.}:

\paragraph{Expectation} compute $p_{ij}$, the probability that Gaussian $j$
generated sample $x^i$.
For the sake of clarity in notations, let us denote
$\mu = \mu_{j}^{(t)}$ and $\Sigma = \Sigma_{j}^{(t)}$ the 
estimated
mean and covariance of Gaussian $j$ at iteration $t$ of the algorithm.
To obtain $p_{ij}$ for a given sample $x^i$ and Gaussian $j$, we first compute the density
\begin{equation}
\label{eq:qij}
q_{ij} = {\cal N}(x^i_o; \mu_o, \Sigma_{oo})
\end{equation}
where
${\cal N}(\cdot; \mu_o, \Sigma_{oo})$ is the Gaussian distribution of mean $\mu_o$
and covariance $\Sigma_{oo}$:
\begin{equation}
\label{eq:gaussian}
{\cal N}(z; \mu_o, \Sigma_{oo}) = \frac{1}{\sqrt{2 \pi |\Sigma_{oo}|^d}} e^{(- \frac{1}{2} (z - \mu_o)^T \Sigma^{-1}_{oo} (z - \mu_o))}.
\end{equation}
$p_{ij}$ is now simply given by
$$
p_{ij} = \frac{q_{ij}}{\sum_{\ell= 1}^L q_{i \ell}}
$$
where $L$ is the total number of Gaussians in the mixture.

\paragraph{Maximization} first fill-in missing values, i.e. define, for each Gaussian~$j$,
$\hat{x}^{i,j}$ by
$\hat{x}^{i,j}_o = x^i_o$ and $\hat{x}^{i,j}_m$ being equal to the expectation 
of the missing values $x^i_m$ given the observed $x^i_o$, assuming Gaussian $j$
has generated $x^i$.
Denoting again $\mu = \mu_{j}^{(t)}$ and $\Sigma = \Sigma_{j}^{(t)}$,
this expectation is equal to
\begin{equation}
\label{eq:xijm}
\hat{x}^{i,j}_m = \mu_{m} + \Sigma_{mo} \Sigma_{oo}^{-1}(x^i_o - \mu_{o}).
\end{equation}
From these $\hat{x}^{i,j}$, the Maximization step of EM yields the new estimates for
the mean and covariances of the Gaussians:
$$
\mu_j^{(t+1)} = \frac{\sum_{i = 1}^n p_{ij} \hat{x}^{i,j}}{\sum_{i = 1}^n p_{ij}}
$$
$$
\Sigma_j^{(t+1)} = \frac{\sum_{i = 1}^n p_{ij} (\hat{x}^{i,j} - \mu_j^{(t+1)}) (\hat{x}^{i,j} - \mu_j^{(t+1)})^T}{\sum_{i = 1}^n p_{ij}} + C_j^{(t)}.
$$
The additional term $C_j^{(t)}$ results from the
imputation of missing values by their conditional expectation, and
is computed as follows (for the sake of clarity,
we denote $C_j^{(t)}$ by $C$ and $\Sigma_j^{(t)}$ by $\Sigma$):
\begin{enumerate}
\item $C$ $\leftarrow$ 0
\item for each $x^i$ with observed and missing parts $x^i_o$, $x^i_m$:
\begin{equation}
C_{mm} \leftarrow C_{mm} + \frac{p_{ij}}{\sum_{k=1}^n p_{kj}} \left( \Sigma_{mm} - \Sigma_{mo} \Sigma_{oo}^{-1} \Sigma_{om} \right)
\label{eq:gaussian:cjt}
\end{equation}
\end{enumerate}
The term being added for each sample corresponds to the covariance
of the missing values $x^i_m$.

Regularization can be added into this framework for instance
by adding a small value to the diagonal of the covariance matrix
$\Sigma_j^{(t)}$, or by keeping only the first $k$ principal components
of this covariance matrix (filling the rest of the data space with
a constant regularization coefficient).

% TODO From reviewer: Sec 2: It would probably be more helpful to see the equation for C_j^(t) rather than to explain it in words.

\section{Scaling EM to Large Datasets}
\label{sec:scaling}

While the EM algorithm naturally extends to problems with missing values,
doing so comes with a high computational cost. As we will show, the
computational burden may be mitigated somewhat by exploiting the similarity
between the covariance matrices between ``nearby'' patterns of missing values.
Before we delve into the details of our algorithm, let us first analyze the
computational costs of the EM algorithm presented above,
for each individual training sample $x^i$:
\begin{itemize}
\item The evaluation of $q_{ij}$ from eq.~\ref{eq:qij} and \ref{eq:gaussian} requires
the inversion of $\Sigma_{oo}$, which costs $O(n_o^3)$ operations,
where $n_o$ is the number of observed variables in $x^i$.
\item The contribution to $C_j^{(t)}$ for each Gaussian $j$ (eq.~\ref{eq:gaussian:cjt}) can be done
in $O(n_o^2 n_m + n_o n_m^2)$ (computation of $\Sigma_{mo} \Sigma_{oo}^{-1} \Sigma_{om}$),
or in $O(n_m^3)$ if $\Sigma^{-1}$ is available, because:
\begin{equation}
\label{eq:cond_covar}
\Sigma_{mm} - \Sigma_{mo} \Sigma_{oo}^{-1} \Sigma_{om} = (\Sigma^{-1})_{mm}^{-1}.
\end{equation}
\end{itemize}

Note that for two examples $x^i$ and $x^k$ with exactly the same pattern of
missing variables, the two expensive operations above need only be
performed once, as they are the same for both $x^i$ and $x^k$.  But in
high-dimensional datasets without a clear structure in the missing values,
most missing patterns are not shared by many samples. For instance, in a
real-world financial dataset we have been working on, the number
of missing patterns is about half the total number of samples.  Since each
iteration of EM has a cost of $O(p L d^3)$, with $p$ unique missing patterns, $L$
components in the mixture, and input dimension $d$, the EM algorithm as
presented in section~\ref{sec:em} is not computationally feasible for large
high-dimensional datasets.
The ``fast'' EM variant proposed in~\citet{Lin+al-2006} also suffers from
the same bottlenecks, i.e. it is fast only when there are few unique missing
patterns.

While the large numbers of unique patterns of missing values typically
found in real-world datasets present a barrier to the application of EM to
these problems, they also motivate a means of reducing the computational
cost. As discussed above, for high-dimensional datasets, the computational
cost is dominated by the determination of the inverse covariance of the observed
variables $\Sigma_{oo}^{-1}$ and of the conditional covariance of the missing
data given the observed data (eq.~\ref{eq:cond_covar}). However, as we will show,
these quantities corresponding to one pattern of the missing values may be
determined from those of another pattern of missing values at a cost
proportional to the distance between the two missing value patterns
(measured as the number of missing and observed variables on which the
patterns differ). Thus, for ``nearby'' patterns of missing values, these
covariance computations may be efficiently computed by chaining their
computation through the set of patterns of missing values.  Furthermore,
since the cost of these updates will be smaller when the missing patterns
of two consecutive samples are close to each other, we want to {\bf
optimize the samples ordering} so as to minimize this cost.

We present the details of the proposed algorithms in the following
sections.  First, we observe in section~\ref{sec:cholesky} how we can avoid
computing $\Sigma_{oo}^{-1}$ by using the Cholesky decomposition of
$\Sigma_{oo}$.  Then we show in section~\ref{sec:ivl} how the so-called
inverse variance lemma can be used to update the conditional covariance
matrix (eq.~\ref{eq:cond_covar}) for a missing pattern given the one computed
for another missing pattern.  As presented in section~\ref{sec:optimal},
these two ideas combined give rise to an objective function with which one
can determine an optimal ordering of the missing patterns, minimizing the
overall computational cost. The resulting fast EM algorithm is summarized
in section~\ref{sec:algo}.

\subsection{Cholesky Updates}
\label{sec:cholesky}

Computing $q_{ij}$ by eq.~\ref{eq:qij} can be done directly from the inverse
covariance matrix $\Sigma_{oo}^{-1}$ as in eq.~\ref{eq:gaussian}, but, as argued
in~\citet{Seeger-2005}, it is just as fast, and numerically more stable,
to use the Cholesky decomposition of
$\Sigma_{oo}$.
Writing $\Sigma_{oo} = L L^T$
with $L$ a lower triangular matrix with positive diagonal elements, we indeed have
$$
z^T \Sigma^{-1}_{oo} z = \| L^{-1} z \|^2
$$
where $L^{-1} z$ can easily be obtained since $L$ is lower triangular.
Assuming $L$ is computed once for the missing pattern of the first sample
in the training set, the question is thus how to update this matrix
for the next missing pattern.
This reduces to finding how to update $L$ when adding or removing rows and columns
to $\Sigma_{oo}$ (adding a row and column when a variable that was missing
is now observed in the next sample, and removing a row and column when a variable that was observed
is now missing).

Algorithms to perform these updates can be found for instance in~\citet{Stewart-1998}.
When adding a row and column, we always add it as the last dimension to minimize the
computations. These are on the order of $O(n_o^2)$, where $n_o$ is the length and
width of $\Sigma_{oo}$.
Removing a row and column is also on the order of $O(n_o^2)$, though the exact cost depends on
the position of the row / column being removed.

Let us denote by $n_d$ the number of differences between two consecutive
missing patterns.
Assuming that $n_d$ is small compared to $n_o$, the above analysis shows that
the overall cost is on the order of $O(n_d \; n_o^2)$ computations.
How to find an ordering of the patterns such that $n_d$ is small will be discussed
in section~\ref{sec:optimal}.

\subsection{Inverse Variance Lemma}
\label{sec:ivl}

%TODO Talk about stability issues.

The second bottleneck of the EM algorithm resides in the computation of eq.~\ref{eq:cond_covar},
corresponding to the conditional covariance of the missing part given the observed part.
Note that we cannot rely on a Cholesky decomposition here, since we need the full
conditional covariance matrix itself.
In order to update $(\Sigma^{-1})_{mm}^{-1}$, we will take advantage of the
so-called inverse variance lemma~\citep{whittaker90}.
It states that the inverse of a partitioned covariance matrix
\begin{equation}
\label{eq:partition}
\Lambda = \left(
\begin{array}{cc}
   \Lambda_{XX} & \Lambda_{XY} \\
   \Lambda_{YX} & \Lambda_{YY}
\end{array}
\right)
\end{equation}
can be computed by
\begin{eqnarray}
\label{eq:ivl}
\Lambda^{-1} =
\left(
\begin{array}{cc}
   \Lambda_{XX}^{-1} + B^T \Lambda_{Y|X}^{-1} B & -B^T \Lambda_{Y|X}^{-1} \\
   - \Lambda_{Y|X}^{-1} B & \Lambda_{Y|X}^{-1}
\end{array}
\right)
\end{eqnarray}
where $\Lambda_{XX}$ is the covariance of the $X$ part,
and the matrix $B$ and the conditional covariance $\Lambda_{Y|X}$ of the $Y$ part given $X$
are obtained by
\begin{equation}
\label{eq:b}
B = \Lambda_{YX} \Lambda_{XX}^{-1}
\end{equation}
\begin{equation}
\label{eq:cond_covar_bis}
\Lambda_{Y|X} = \Lambda_{YY} - \Lambda_{YX} \Lambda_{XX}^{-1} \Lambda_{XY}
\end{equation}
Note that eq.~\ref{eq:cond_covar_bis} is similar to eq.~\ref{eq:cond_covar},
since the conditional covariance of $Y$ given $X$ verifies
\begin{equation}
\label{eq:cond_identity}
\Lambda_{Y|X} = (\Lambda^{-1})_{YY}^{-1}
\end{equation}
where we have also partitioned the inverse covariance matrix as
\begin{equation}
\label{eq:partition_inv}
\Lambda^{-1} = \left(
\begin{array}{cc}
   (\Lambda^{-1})_{XX} & (\Lambda^{-1})_{XY} \\
   (\Lambda^{-1})_{YX} & (\Lambda^{-1})_{YY}
\end{array}
\right).
\end{equation}

These equations can be used to update the conditional covariance matrix of
the missing variables given the observed variables when going from one
missing pattern to the next one, so that it does not need to be re-computed
from scratch.  Let us first consider the case of going from sample $x^i$ to
$x^j$, where we only add missing values (i.e.  all variables that are
missing in $x^i$ are also missing in $x^j$).  We can apply the inverse
variance lemma (eq.~\ref{eq:ivl}) with the following quantities:
\begin{itemize}
\item $\Lambda^{-1}$ is the conditional covariance\footnote{Note that in
    the original formulation of the inverse variance lemma $\Lambda$ is a
    covariance matrix, while we use it here as an inverse covariance: since
    the inverse of a symmetric positive definite matrix is also symmetric
    positive definite, it is possible to apply eq.~\ref{eq:ivl} to an inverse
    covariance.  } of the missing variables in $x^j$ given the observed
  variables, i.e. $(\Sigma^{-1})_{mm}^{-1}$ in eq.~\ref{eq:cond_covar}, the
  quantity we want to compute,
\item $X$ are the missing variables in sample $x^i$,
\item $Y$ are the missing variables in sample $x^j$ that were not missing in $x^i$,
\item $\Lambda_{XX}^{-1}$ is the conditional covariance of the missing variables in $x^i$ given
the observed variables, that would have been computed previously,
\item since $\Lambda = (\Sigma^{-1})_{mm}$, then $\Lambda_{YX}$ and $\Lambda_{YY}$ are simply sub-matrices
of the global inverse covariance matrix, that only needs to be computed once (per iteration
of EM).
\end{itemize}

Let us denote by $n_d$ the number of missing values added when going from $x^i$ to $x^j$,
and by $n_m$ the number of missing values in $x^i$. Assuming $n_d$ is small compared to
$n_m$, then the computational cost of eq.~\ref{eq:ivl} is dominated by the cost
$O(n_d \; n_m^2)$ for the computation of $B$ by eq.~\ref{eq:b} and of the upper-left term
in eq.~\ref{eq:ivl} (the inversion of $\Lambda_{Y|X}$ is only in $O(n_d^3)$).

In the case where we remove missing values instead of adding some, this corresponds
to computing $\Lambda_{XX}^{-1}$ from $\Lambda^{-1}$ using eq.~\ref{eq:ivl}.
This can be done from the partition of $\Lambda^{-1}$ since, by identifying eq.~\ref{eq:ivl}
and \ref{eq:partition_inv}, we have:
\begin{eqnarray*}
&& (\Lambda^{-1})_{XX} - (\Lambda^{-1})_{XY} (\Lambda^{-1})_{YY}^{-1} (\Lambda^{-1})_{YX} \\
&=&
\Lambda_{XX}^{-1} + B^T \Lambda_{Y|X}^{-1} B - B^T \Lambda_{Y|X}^{-1} (\Lambda^{-1})_{YY}^{-1} \Lambda_{Y|X}^{-1} B \\
&=& \Lambda_{XX}^{-1}
\end{eqnarray*}
where we have used eq.~\ref{eq:cond_identity} to obtain the final result.
Once again the cost of this computation is dominated by a term of the form
$O(n_d \; n_m^2)$, where this time $n_d$ denotes the number of missing values
that are removed when going from $x^i$ to $x^j$ and $n_m$ the number of missing values
in $x^j$.

Thus, in the general case where we both remove and add missing values, the cost
of the update is on the order of $O(n_d \; n_m^2)$, if we denote by $n_d$ the total
number of differences in the missing patterns, and by $n_m$ the average number
of missing values in $x^i$ and $x^j$ (which are assumed to be close, since $n_d$
is supposed to be small).
The speed-up is on the order of $O(n_m / n_d)$ compared to the ``naive'' algorithm that would
re-compute the conditional covariance matrix for each missing pattern.

\subsection{Optimal Ordering From the Minimum Spanning Tree}
\label{sec:optimal}

Given an ordering $\{m^1, m^2, \ldots, m^p\}$ of the $p$ missing patterns
present in the training set, during an iteration of the EM algorithm we have to:
\begin{enumerate}
\item Compute the Cholesky decomposition of $\Sigma_{oo}$ and the conditional
covariance $(\Sigma^{-1})_{mm}^{-1}$ for the first missing pattern $m^1$.
\item For each subsequent missing pattern $m^i$, find the
missing pattern in $\{m^1, m^2, \ldots, m^{i-1}\}$ that allows the fastest computation
of the same matrices, from the update methods presented in sections~\ref{sec:cholesky}
and~\ref{sec:ivl}.
\end{enumerate}

Since each missing pattern but the first one has a ``parent'' (the missing pattern
from which we update the desired matrices), we visit the missing patterns
in a tree-like fashion: {\em the optimal tree is thus the minimum spanning tree
of the fully-connected graph whose nodes are the missing patterns and the weight
between nodes $m^i$ and $m^j$ is the cost of computing matrices for $m^j$ given
matrices for $m^i$}.
Note that the spanning tree obtained this way is the exact optimal solution and not
an approximation (assuming we constrain ourselves to visiting only observed missing patterns
and we can compute the true cost of updating matrices).

For the sake of simplicity, we used the number of differences $n_d$ between
missing pattterns $m^i$ and $m^j$ as the weights between two nodes. Finding the
``true'' cost is difficult because (1)~it is implementation-dependent and (2)~it
depends on the ordering of the columns for the Cholesky updates, and this
ordering varies depending on previous updates
due to the fact it is more efficient to add new dimensions as the last ones,
as argued in section~\ref{sec:cholesky}.
We tried more sophisticated variants of the cost function, but they
did not decrease significantly the overall computation time.

Note it would be possible to allow the creation of ``virtual'' missing
patterns, whose corresponding matrices could be used by multiple observed
missing patterns in order to speed-up updates even further.  Finding the
optimal tree in this setting corresponds to finding the optimal Steiner tree~\citep{Hwang+al-1992},
which is known to be NP-hard.  Since we do not expect
the available approximate solution schemes to provide a huge speed-up, we
did not explore this approach further.

Finally, one may have concerns about the numerical stability of this approach,
since computations are incremental and thus numerical errors will be accumulated.
The number of incremental steps is directly linked to the depth of the minimum
spanning tree, which will often be logarithmic in the number of training samples,
but may grow linearly in the worst case.
Although we did not face this problem in our own experiments (where the accumulation
of errors never led to results significantly different from the exact solution),
the following heuristics can be used to solve it:
the matrices of interest can be re-computed ``from scratch'' at each node of the
tree whose depth is a multiple of $k$, with $k$ a hyper-parameter trading accuracy
for speed.

\subsection{Fast EM Algorithm Overview}
\label{sec:algo}

We summarize here the previous sections by giving a sketch of the resulting fast
EM algorithm for Gaussian mixtures:
\begin{enumerate}
\item Find all unique missing patterns in the dataset.
\item Compute the minimum spanning tree\footnote{Since the cost of this computation is
in $O(p^2)$, with $p$ the number of missing patterns, if $p$ is too large it is possible to perform an initial
basic clustering of the missing patterns and compute the minimum spanning tree independently in
each cluster.}
 of the corresponding graph of missing patterns (see section~\ref{sec:optimal}).
\item Deduce from the minimum spanning tree on missing patterns an ordering of the training samples (since each
missing pattern may be common to many samples). \label{step:mst}
\item Initialize the means of the mixture components by the $K$-means clustering algorithm,
and their covariances from the empirical covariances in each cluster (either imputing missing values
with the cluster means or just ignoring them, which is what we did in our experiments).
\item Iterate through EM steps (as described in section~\ref{sec:em}) until convergence (or until
a validation error increases). At each step, the expensive matrix computations highlighted
in section~\ref{sec:scaling} are sped-up by using iterative updates, following the ordering
obtained in step~\ref{step:mst}.
\end{enumerate}

\section{Experiments}
\label{sec:experiments}

% TODO
%However, the analysis does not compare run-times, in particular between a full inversion and an interative inversion (like the one presented here), although the iterative inversion of \Sigma_{oo} is one of the main theoretical results of the paper.

% The paper is well written and structured. A minor remark would be to add axis labels in Figure 1 and 2 and enlarge the fonts for better readability.

\subsection{Learning to Model Images}

To assess the speed improvement of our proposed algorithm over the ``naive'' EM
algorithm, we trained mixtures of Gaussians on the MNIST dataset of
handwritten digits.
For each class of digits (from 0 to 9), we optimized an individual mixture
of Gaussians in order to model the class distribution.
We manually added missing values by removing the pixel information in each image from
a randomly chosen square of 5x5 pixels (the images are 28x28 pixels, i.e. in dimension 784).
The mixtures were first trained efficiently on the first 4500 samples of each class, while the
rest of the samples were used to select the hyperparameters, namely the number
of Gaussians (from 1 to 10), the fraction of principal components kept (75\%,
90\% or 100\%), and the random number generator seed used in the mean initialization
(chosen between 5 different values).
The best model was chosen based on the average negative log-likelihood.
It was then re-trained using the ``naive'' version of the EM algorithm, in
order to compare execution time and also ensure the same results were obtained.

On average, the speed-up on our cluster computers (32 bit P4 3.2~Ghz with 2~Gb of memory)
was on the order of 8.
We also observed a larger improvement (on the order of 20) on another
architecture (64 bit Athlon 2.2~Ghz with 2~Gb of memory): the difference seemed to be due
to implementations of the BLAS and LAPACK linear algebra libraries.

\begin{figure}[htbp]
\begin{center}
\resizebox{0.18\textwidth}{!}{\includegraphics{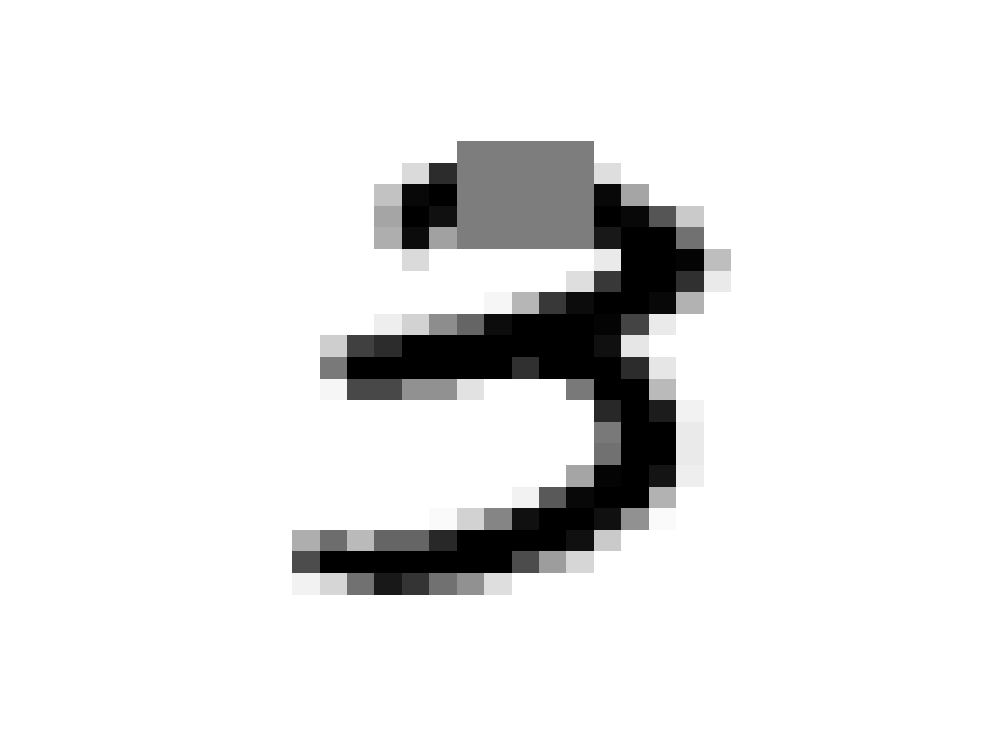}}
\resizebox{0.18\textwidth}{!}{\includegraphics{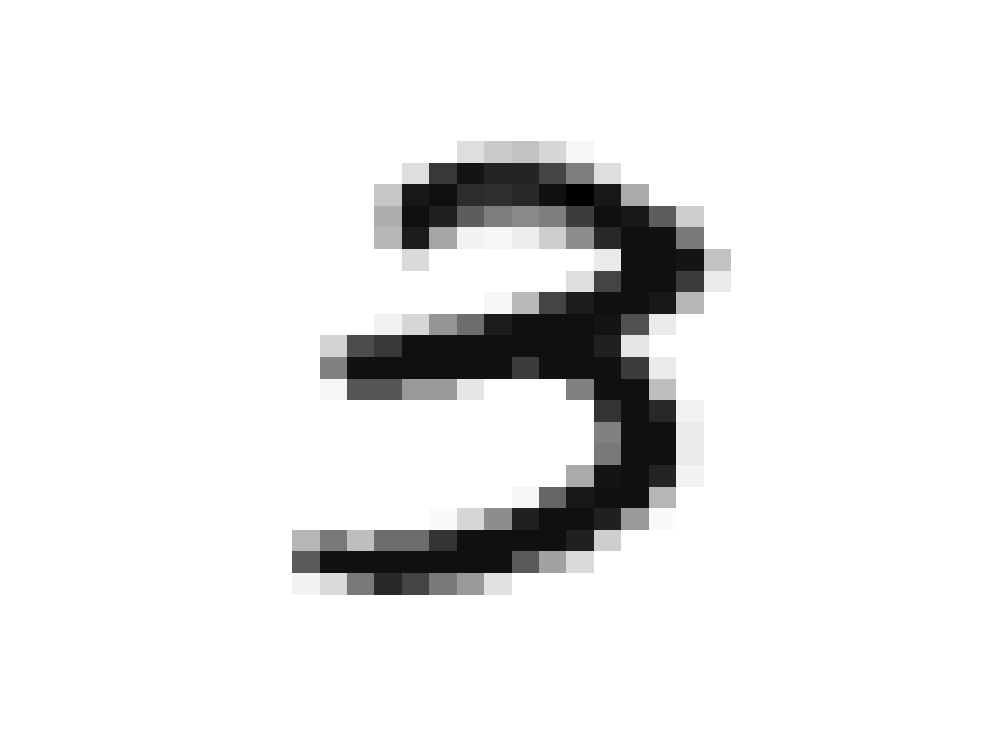}}\\
\resizebox{0.18\textwidth}{!}{\includegraphics{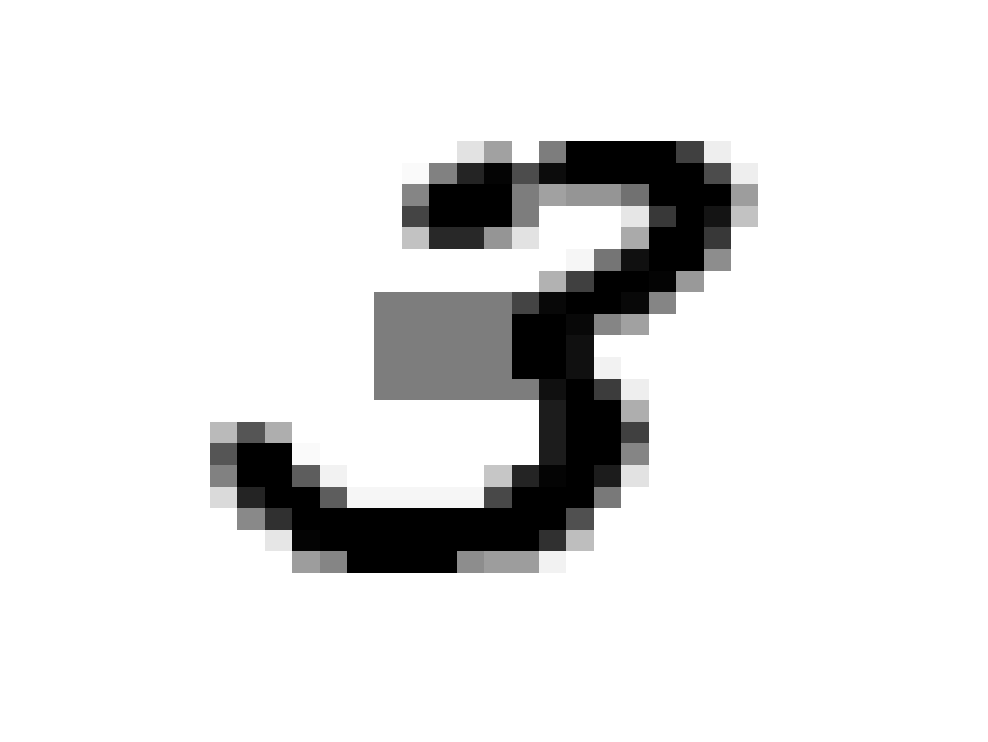}}
\resizebox{0.18\textwidth}{!}{\includegraphics{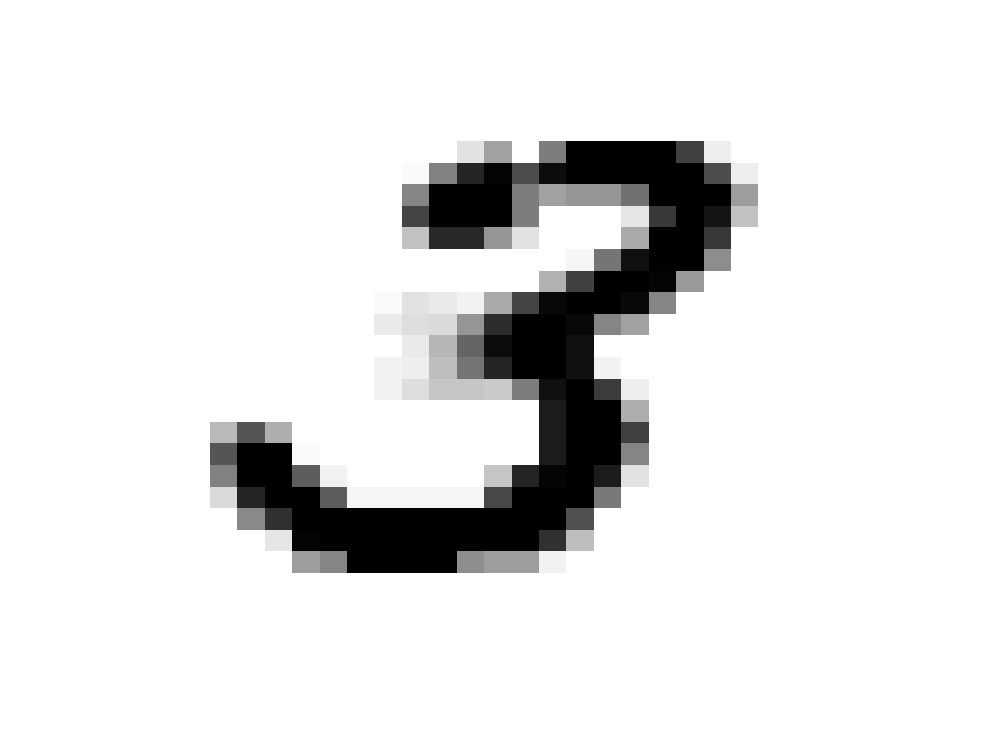}}
\end{center}
\caption{
Imputation of missing pixels in images.
On the left, images with missing pixels (represented by a grey square).
On the right, the same images with the missing pixels imputed by a
mixture of Gaussians.}
\label{fig:examples}
\end{figure}

We display in figure~\ref{fig:examples} the imputation of missing values realized by the
trained mixture when provided with sample test images.
On each row, images with grey squares have missing values (identified by these squares),
while images next to them show the result of the missing value imputation.
Although the imputed pixels are a bit fuzzy, the figure shows
the mixture was able to capture meaningful correlations between pixels,
and to impute sensible missing values.

\subsection{Combining Generative and Discriminative Models}

The Abalone dataset from the UCI Machine Learning Repository is a standard
benchmark regression task. The official training set (3133 samples) is divided
into a training (2000) and validation set (1133), while we use the official
test set (1044).
This dataset does not contain any missing data, which allows us to see how the
algorithms behave as we add more missing values. We systematically preprocess the dataset
after inserting missing values, by normalizing all variables (including the target)
so that the mean and standard deviation on the training set are respectively 0 and 1
(note that we do not introduce missing values in the target, so that mean squared errors
can be compared).

We compare three different missing values imputation mechanisms:
\begin{enumerate}
\item Imputation by the conditional expectation of the missing values
as computed by a mixture of Gaussians
learnt on the joint distribution of the input and target (the algorithm proposed
in this paper)
\item Imputation by the global empirical mean (on the training set)
\item Imputation by the value found in the nearest neighbor
that has a non missing value for this variable (or, alternatively, by the mean
of the 10 such nearest neighbors). Because there is no obvious way to
compute the nearest neighbors in the presence of missing values (see e.g.~\citep{Caruana-2001}),
we allow this algorithm to compute the neighborhoods
based on the original dataset with no missing value: it is thus expected
to give the optimal performance that one could obtain with such a nearest-neighbor
algorithm.
\end{enumerate}

One one hand, we report the performance of the mixture of Gaussian used directly as a predictor for regression.
On another hand, the imputed values are also fed to the two following discriminant algorithms, whose hyper-parameters
are optimized on the validation set:
\begin{enumerate}
\item A one-hidden-layer feedforward neural network trained by stochastic gradient
descent, with hyper-parameters the number of hidden units (among 5, 10, 15, 20, 30, 50, 100, 200),
the quadratic weight decay (among 0, $10^{-6}$, $10^{-4}$, $10^{-2}$), the initial learning rate (among $10^{-2}$, $10^{-3}$, $10^{-4}$)
and its decrease constant\footnote{The learning rate after seeing $t$ samples is equal
to $\mu(t) = \frac{\mu(0)}{1 + \lambda t}$, where $\mu(0)$ is the initial learning rate
and $\lambda$ the decrease constant.} (among 0, $10^{-2}$, $10^{-4}$, $10^{-6}$).
\item A kernel ridge regressor, with hyper-parameters the weight decay (in $10^{-8}$, $10^{-6}$, $10^{-4}$, $10^{-2}$, 1)
and the kernel: either the linear kernel, the Gaussian kernel (with bandwidth in 100, 50, 10, 5, 1, 0.5, 0.1, 0.05, 0.01)
or the polynomial kernel (with degree in 1, 2, 3, 4, 5 and dot product scaling coefficient in 0.01, 0.05, 0.1, 0.5, 1, 5, 10).
\end{enumerate}

\begin{center}
\begin{figure}[htbp]
\resizebox{0.4\textwidth}{!}{\includegraphics{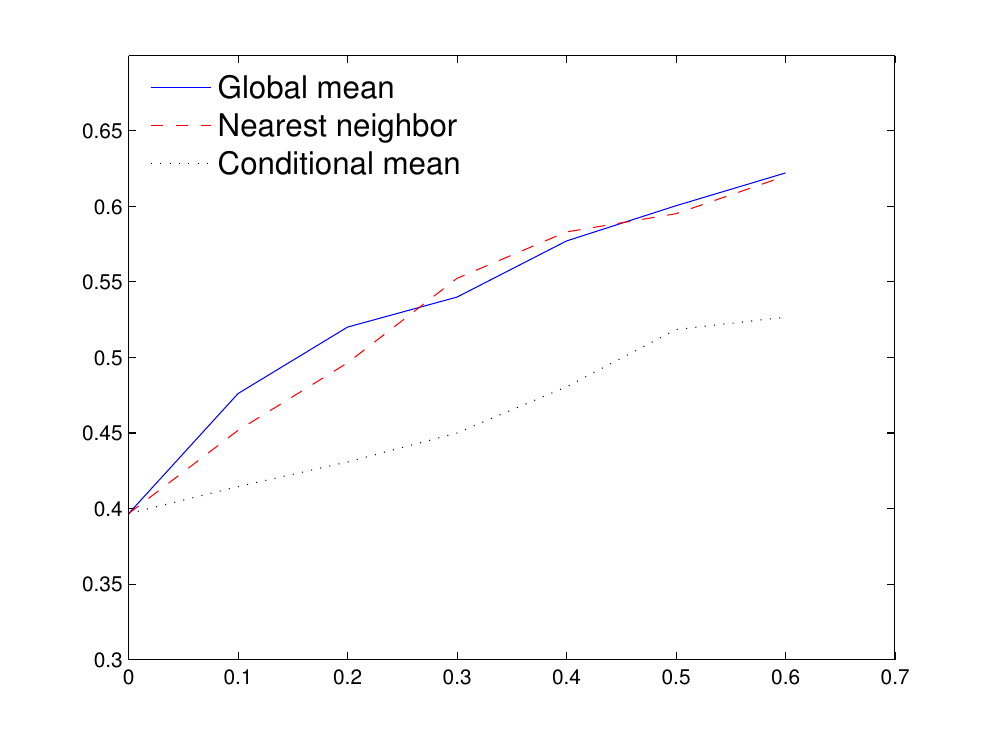}}
\resizebox{0.4\textwidth}{!}{\includegraphics{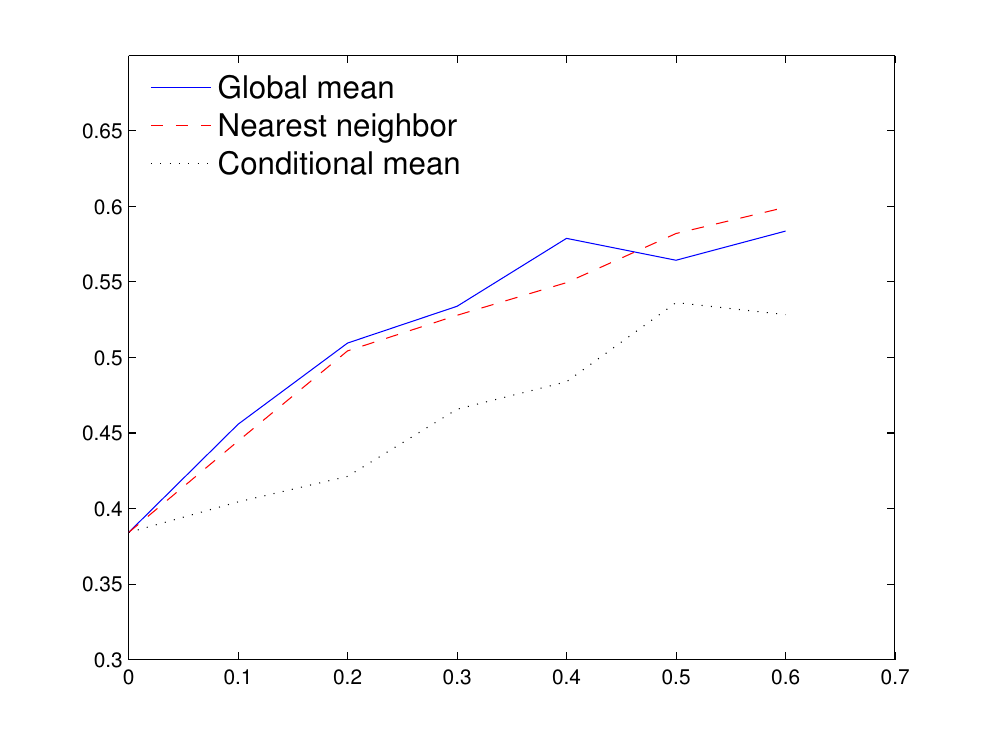}}
\caption{
Test mean-squared error ($y$ axis) on Abalone when the proportion of missing values increases ($x$ axis).
The three missing values imputation mechanisms are compared when
using a neural network (top), and kernel ridge regression (bottom).}
\label{fig:imput_compare}
\end{figure}
\end{center}

Figure~\ref{fig:imput_compare} compares the three missing values imputation mechanisms
when using a neural network and kernel ridge regression.
It can be seen that the conditional mean imputation obtained by the Gaussian mixture
significantly outperforms the global mean imputation and nearest neighbor imputation (which
is tried with both 1 and 10 neighbors, keeping the best on the validation set).
The latter seems to be reliable only when there are few missing values in the dataset: this
is expected, as when the number of missing values increases
one has to go further in space to find neighbors that contain non-missing
values for the desired variables.

\begin{center}
\begin{figure}[htbp]
\resizebox{0.4\textwidth}{!}{\includegraphics{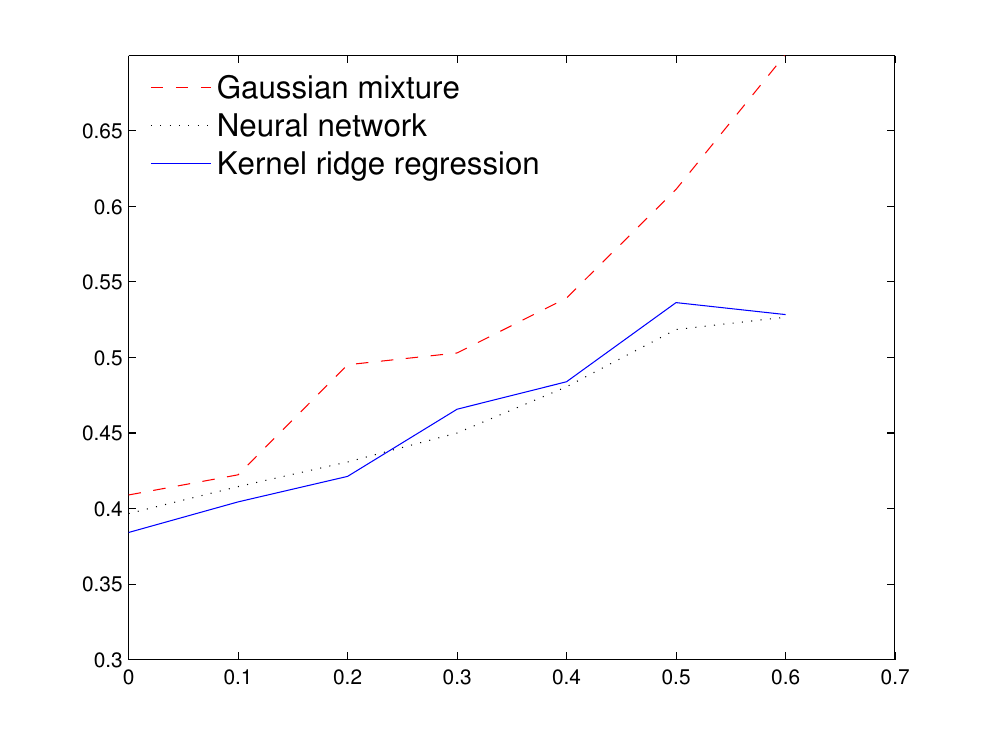}}
\caption{
Test mean-squared error ($y$ axis) on Abalone when the proportion of missing values increases ($x$ axis).
Combining a discriminant algorithm with the generative Gaussian mixture model
works better than the Gaussian mixture alone
(both the neural network and the kernel ridge
regressor use here the conditional mean imputation of missing values provided
by the Gaussian mixture).}
\label{fig:discriminant}
\end{figure}
\end{center}

Figure~\ref{fig:discriminant} illustrates the gain of combining the generative
model (the mixture of Gaussian) with the discriminant learning algorithms:
even though the mixture can be used directly as a regressor (as argued
in the introduction), its prediction
accuracy can be greatly improved by a supervised learning step.

\section{Conclusion}
\label{sec:conclusion}

In this paper, we considered the problem of training Gaussian mixtures in the context of
large high-dimensional datasets with a significant fraction
of the data matrix missing.
In such situations, the application of
EM to the imputation of missing values results in expensive matrix
computations. We have proposed a more efficient algorithm that uses
matrix updates over a minimum spanning tree of missing patterns
to speed-up these matrix computations, by an order of magnitude.

We also explored the application of a hybrid
scheme where a mixture of Gaussians generative model, trained with EM, is
used to impute the missing values with their conditional means. These
imputed datasets were then used in a discriminant learning model (neural
networks and kernel ridge regression) where they were shown to provide
significant improvement over more basic missing value
imputation methods.

% use section* for acknowledgement
%\section*{Acknowledgment}

%The authors would like to thank...

% Can use something like this to put references on a page
% by themselves when using endfloat and the captionsoff option.
%\ifCLASSOPTIONcaptionsoff
%  \newpage
%\fi

% trigger a \newpage just before the given reference
% number - used to balance the columns on the last page
% adjust value as needed - may need to be readjusted if
% the document is modified later
%\IEEEtriggeratref{8}
% The "triggered" command can be changed if desired:
%\IEEEtriggercmd{\enlargethispage{-5in}}

% references section

% can use a bibliography generated by BibTeX as a .bbl file
% BibTeX documentation can be easily obtained at:
% http://www.ctan.org/tex-archive/biblio/bibtex/contrib/doc/
% The IEEEtran BibTeX style support page is at:
% http://www.michaelshell.org/tex/ieeetran/bibtex/
\bibliographystyle{IEEEtran}

\begin{thebibliography}{12}

% Generated by IEEEtran.bst, version: 1.13 (2008/09/30)
\providecommand{\url}[1]{#1}
\csname url@samestyle\endcsname
\providecommand{\newblock}{\relax}
\providecommand{\bibinfo}[2]{#2}
\providecommand{\BIBentrySTDinterwordspacing}{\spaceskip=0pt\relax}
\providecommand{\BIBentryALTinterwordstretchfactor}{4}
\providecommand{\BIBentryALTinterwordspacing}{\spaceskip=\fontdimen2\font plus
\BIBentryALTinterwordstretchfactor\fontdimen3\font minus
  \fontdimen4\font\relax}
\providecommand{\BIBforeignlanguage}[2]{{%
\expandafter\ifx\csname l@#1\endcsname\relax
\typeout{** WARNING: IEEEtran.bst: No hyphenation pattern has been}%
\typeout{** loaded for the language `#1'. Using the pattern for}%
\typeout{** the default language instead.}%
\else
\language=\csname l@#1\endcsname
\fi
#2}}
\providecommand{\BIBdecl}{\relax}
\BIBdecl

\bibitem{Parzen62}
E.~Parzen, ``On the estimation of a probability density function and mode,''
  \emph{Annals of Mathematical Statistics}, vol.~33, pp. 1064--1076, 1962.

\bibitem{Dempster77}
A.~P. Dempster, N.~M. Laird, and D.~B. Rubin, ``Maximum-likelihood from
  incomplete data via the {EM} algorithm,'' \emph{Journal of Royal Statistical
  Society B}, vol.~39, pp. 1--38, 1977.

\bibitem{Zoubin-nips94}
Z.~Ghahramani and M.~I. Jordan, ``Supervised learning from incomplete data via
  an {EM} approach,'' in \emph{Advances in Neural Information Processing
  Systems 6 (NIPS'93)}, D.~Cowan, G.~Tesauro, and J.~Alspector, Eds.\hskip 1em
  plus 0.5em minus 0.4em\relax San Mateo, CA: Morgan Kaufmann, 1994.

\bibitem{Bahl86}
L.~Bahl, P.~Brown, P.~{deSouza}, and R.~Mercer, ``Maximum mutual information
  estimation of hidden markov parameters for speech recognition,'' in
  \emph{International Conference on Acoustics, Speech and Signal Processing
  (ICASSP)}, Tokyo, Japan, 1986, pp. 49--52.

\bibitem{Lin+al-2006}
T.~I. Lin, J.~C. Lee, and H.~J. Ho, ``On fast supervised learning for normal
  mixture models with missing information,'' \emph{Pattern Recognition},
  vol.~39, no.~6, pp. 1177--1187, Jun. 2006.

\bibitem{DiZio+al-2007}
M.~D. Zio, U.~Guarnera, and O.~Luzi, ``Imputation through finite {G}aussian
  mixture models,'' \emph{Computational Statistics \& Data Analysis}, vol.~51,
  no.~11, pp. 5305--5316, 2007.

\bibitem{Little+Rubin-2002}
R.~J.~A. Little and D.~B. Rubin, \emph{Statistical Analysis with Missing Data},
  2nd~ed.\hskip 1em plus 0.5em minus 0.4em\relax New York: Wiley, 2002.

\bibitem{Seeger-2005}
M.~Seeger, ``Low rank updates for the {Cholesky} decomposition,'' Department of
  EECS, University of California at Berkeley, Tech. Rep., 2005.

\bibitem{Stewart-1998}
G.~W. Stewart, \emph{Matrix Algorithms, Volume {I}: Basic
  Decompositions}.\hskip 1em plus 0.5em minus 0.4em\relax Philadelphia: SIAM,
  1998.

\bibitem{whittaker90}
J.~Whittaker, \emph{Graphical Models in Applied Multivariate Statistics}.\hskip
  1em plus 0.5em minus 0.4em\relax Wiley, Chichester, 1990.

\bibitem{Hwang+al-1992}
F.~K. Hwang, D.~Richards, and P.~Winter, ``The {Steiner} tree problem,''
  \emph{Annals of Discrete Mathematics}, vol.~53, 1992.

\bibitem{Caruana-2001}
R.~Caruana, ``A non-parametric {EM}-style algorithm for imputing missing
  values,'' in \emph{Proceedings of the Eigth International Workshop on
  Artificial Intelligence and Statistics (AISTATS'01)}.\hskip 1em plus 0.5em
  minus 0.4em\relax Society for Artificial Intelligence and Statistics, 2001.

\end{thebibliography}
% <OR> manually copy in the resultant .bbl file
% set second argument of \begin to the number of references
% (used to reserve space for the reference number labels box)

% biography section
% 
% If you have an EPS/PDF photo (graphicx package needed) extra braces are
% needed around the contents of the optional argument to biography to prevent
% the LaTeX parser from getting confused when it sees the complicated
% \includegraphics command within an optional argument. (You could create
% your own custom macro containing the \includegraphics command to make things
% simpler here.)
%\begin{biography}[{\includegraphics[width=1in,height=1.25in,clip,keepaspectratio]{mshell}}]{Michael Shell}
% or if you just want to reserve a space for a photo:

%\begin{IEEEbiography}{Michael Shell}
%Biography text here.
%\end{IEEEbiography}

% if you will not have a photo at all:
%\begin{IEEEbiographynophoto}{John Doe}
%Biography text here.
%\end{IEEEbiographynophoto}

% insert where needed to balance the two columns on the last page with
% biographies
%\newpage

%\begin{IEEEbiographynophoto}{Jane Doe}
%Biography text here.
%\end{IEEEbiographynophoto}

% You can push biographies down or up by placing
% a \vfill before or after them. The appropriate
% use of \vfill depends on what kind of text is
% on the last page and whether or not the columns
% are being equalized.

%\vfill

% Can be used to pull up biographies so that the bottom of the last one
% is flush with the other column.
%\enlargethispage{-5in}

% that's all folks
\end{document}